\documentclass[11pt, a4paper, gdm]{google}

\usepackage[authoryear, sort&compress, round]{natbib}
\usepackage{tikz}
\usetikzlibrary{shapes, arrows, positioning}
\usepackage{subcaption}
\usepackage{listings}
\usepackage{lineno}
\newcommand{\hidecaption}{\phantomcaption}

\uselogo{}

\title{Analysis of Optimality of Large Language Models on Planning Problems}

\renewcommand\Affilfont{\normalfont\fontsize{8}{10}\selectfont}

\makeatletter
\renewcommand\AB@authnote[1]{}
\renewcommand\AB@affilnote[1]{}
\makeatother
\author{Bernd Bohnet, Michael C. Mozer, Kevin Swersky, Wil Cunningham,\\
Aaron Parisi, Kathleen Kenealy and Noah Fiedel}
\affil{Google DeepMind}

\begin{abstract}
Classic AI planning problems have been revisited in the Large Language Model (LLM) era, with a focus
of recent benchmarks on success rates rather than plan efficiency. We examine the degree to 
which frontier models reason \emph{optimally} versus relying on simple, heuristic, and possibly inefficient 
strategies. We focus on the Blocksworld domain involving towers of labeled blocks which have to be
moved from an initial to a goal configuration via a set of primitive actions. We also study a formally
equivalent task, the generalized Path-Star ($P^*$) graph, in order to isolate true topological reasoning 
from semantic priors. We systematically manipulate problem depth (the height of block towers), width
(the number of towers), and compositionality (the number of goal blocks).
Reasoning-enhanced LLMs significantly outperform traditional satisficing planners (e.g., LAMA) in complex, multi-goal configurations. Although classical search algorithms hit a wall as the search space
expands, LLMs track theoretical optimality limits with near-perfect precision, even when domain-specific semantic hints are stripped away. To explain these surprising findings, we consider (and find evidence to support) two hypotheses: an active Algorithmic Simulation executed via reasoning tokens and a Geometric Memory that allows models to represent the $P^*$ topology as a navigable global geometry, effectively bypassing exponential combinatorial complexity.
\end{abstract}

\begin{document}

\maketitle

\section{Introduction}

The capability of Large Language Models (LLMs) to perform autonomous planning has been a subject of intense debate and rapid evolution \citep{lehnert2024beyond,valmeekam2023planningabilitieslargelanguage, zhao2025improvinglargelanguagemodel,jobs2025blocksworld,bohnet2025enhancingllmplanningcapabilities}. While early investigations suggested that LLMs fundamentally lacked the reasoning capabilities required for planning, famously characterized by \citet{valmeekam2023planningabilitieslargelanguage} as a domain where models struggle to generalize, the frontier of generative AI has shifted dramatically. Recent evaluations indicate that the latest generation of models has transcended simple puzzle-solving to handle substantial combinatorial complexity. Notably, \citet{correa2025planning} demonstrated that frontier models such as GPT-5 and Gemini 2.5 Pro have achieved parity with robust classical planners like LAMA \citep{richter2010lama,richter2011lama} on standard benchmarks. Their work confirms that these models can generate valid plans for Blocksworld tasks involving up to 477 blocks and Rovers tasks with 29 agents, with valid solution paths reaching 1,194 steps in length. These results effectively reduce the performance gap between neural and symbolic planning systems on standard PDDL tasks.

\begin{figure*}[t] 
    \centering
    \begin{subfigure}[c]{0.45\textwidth}
        \centering
        \vspace{0pt}
        \includegraphics[width=\textwidth]{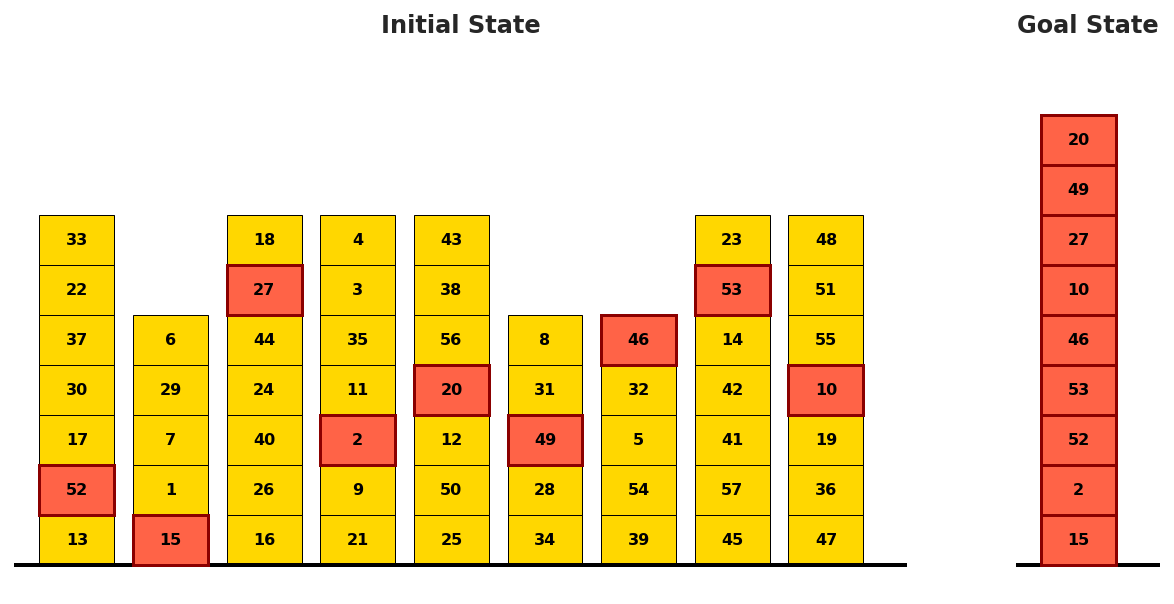} 
        \label{fig:graphA}
    \end{subfigure}
    \hfill 
    \begin{subfigure}[c]{0.54\textwidth}
        \centering
        \vspace{0pt}
        \includegraphics[width=\textwidth]{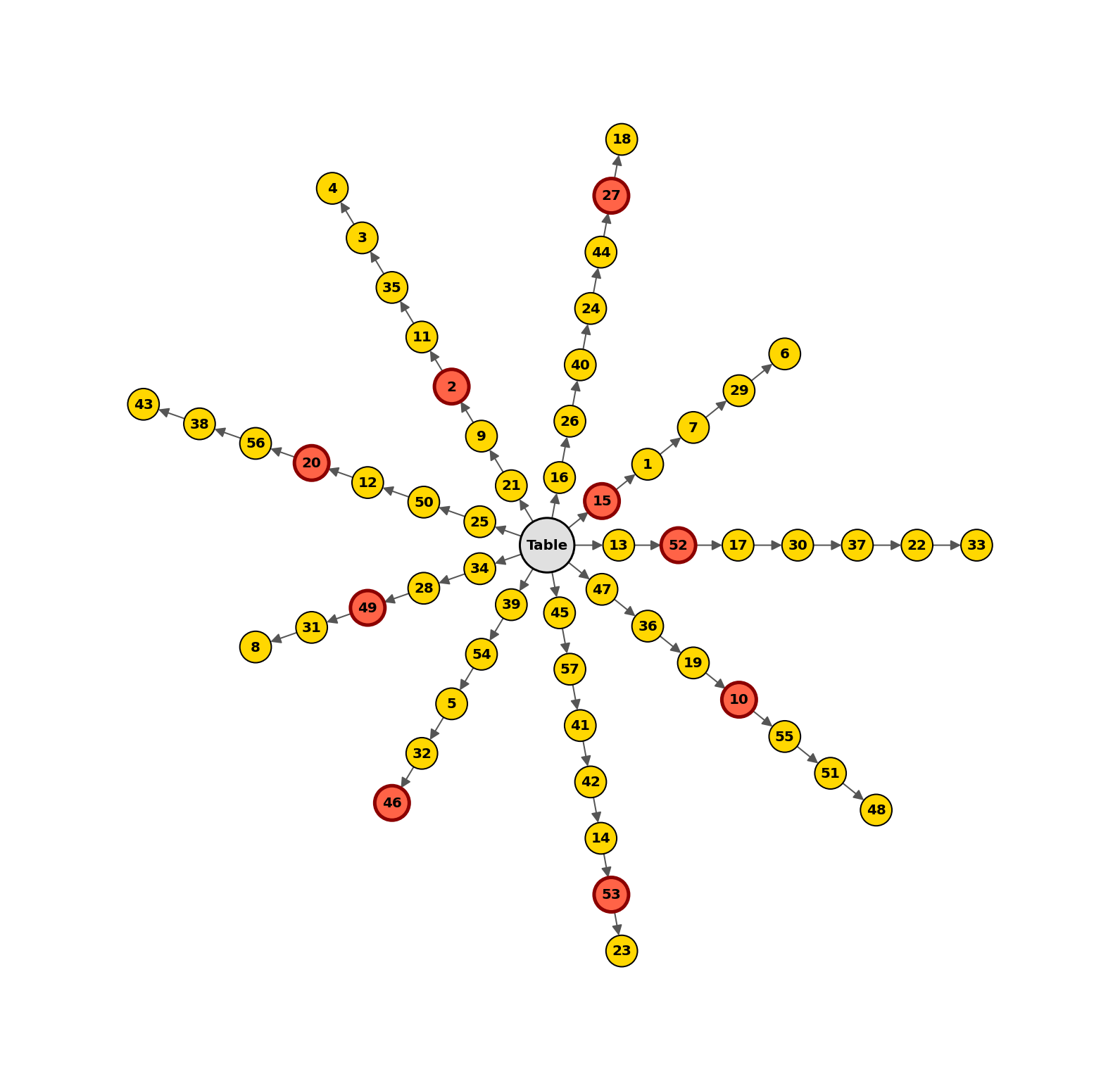}
        \label{fig:graphB}
    \end{subfigure}
    \vspace{-1.5em}
    \caption{\textbf{Structural Isomorphism between Blocksworld and $P^*$ Topology.} \textbf{(a)} A standard Blocksworld instance illustrating the initial state (left) and the target goal state (right). The task requires rearranging blocks using four standard atomic actions: unstack, stack, pick-up, and put-down. \textbf{(b)} The corresponding $P^*$ graph representation of the initial state. The table serves as the central root node, and each stack of blocks forms a disjoint branch, framing the planning task as a structural path-traversal problem.}
    \label{fig:both_graphs}    
\end{figure*}

However, while coverage metrics on standard benchmarks prove that LLMs \textit{can} plan, they often obscure \textit{how} these models reason and where their boundaries lie regarding plan quality and structural complexity. In this work, we move beyond binary success/failure metrics to analyze the \textbf{optimality} and \textbf{structural limits} of LLM planners by framing the problem through a topological lens. Specifically, we posit that the reasoning required for domains like Blocksworld is structurally isomorphic to a generalized $P^*$ (``Path-Star'') graph topology \citep{bachmann24pitfalls}. In this formulation, the ``Table'' acts as a central root node, and each vertical stack of blocks constitutes a disjoint branch. Solving the problem becomes equivalent to solving a series of dynamic $P^*$ traversals: the model must identify the correct branch containing a goal block, trace the dependency chain upwards, and execute a top-down unstacking plan. By establishing this structural isomorphism, we can rigorously investigate whether LLMs rely on local heuristic associations or engage in systematic multihop traversal of the problem's dependency structure. To cleanly isolate this capability without semantic confounding factors, we restrict our experimental scope to configurations where each stack houses a single goal block or none.

To investigate this, we constructed a space of distinct scenarios, all rooted in this p-star problem structure. These scenarios range from a ``Grand Challenge'' where problem volume scales holistically, to targeted regimes that isolate specific complexities such as building increasingly higher towers (scaling depth), managing a massive number of towers (scaling width), or resolving dense chains of increasing goals (scaling composition).

%
Unlike previous studies that rely on complex prompting strategies to boost performance, we focus on the intrinsic reasoning capabilities of the models. We employ a minimal prompting strategy, providing only the PDDL domain, the problem definition, and a single example, forcing the model to rely on pure inference rather than improving in-context learning.

To understand the reasoning demands of the Blocksworld domain, we formally map it to the "Path-Star" ($P^*$) topology identified by \citet{bachmann24pitfalls} 
as a critical stress test for deep sequence models. The $P^*$ graph consists of a central root node with multiple disjoint paths (branches) radiating outwards, see Figure~\ref{fig:both_graphs} (right). In the context of sequence modeling, the "hardest token" to predict is determining the correct branch to traverse, as it requires the model to implicitly trace a path back from a distant goal leaf to the root.

We posit that a Blocksworld state is structurally isomorphic to a generalized $P^*$ graph where the ``Table'' acts as the central root ($v_{root}$) and each vertical stack of blocks constitutes a disjoint branch (Figure~\ref{fig:both_graphs}b). Solving a Blocksworld problem is, therefore, equivalent to solving a series of dynamic $P^*$ traversals. Given a goal configuration such as the target tower shown in Figure~\ref{fig:both_graphs}a (right), the planner must reconstruct it from the bottom up. For each goal block $b_{goal}$, starting with the lowest in the target tower and proceeding upwards, the procedure is: \textbf{(1) Branch Identification:} Among all towers in the initial state, locate the specific branch containing $b_{goal}$. Since the problem is presented as a shuffled list of predicates, this requires searching the full context to resolve which tower holds $b_{goal}$. \textbf{(2) Iterative Clearing:} Starting from the top of the identified branch, repeatedly move the topmost block to the table until $b_{goal}$ is exposed. Each removal requires two atomic actions (unstack and put-down), yielding a clearing cost of $2 \cdot \text{depth}(b_{goal})$. \textbf{(3) Goal Placement:} Once exposed, move $b_{goal}$ to its specified position in the goal tower (pick-up and stack). This three-step cycle is repeated for every target block in sequence, constructing the goal tower from its base upwards.

To illustrate the difficulty of Branch Identification, consider a small example with three towers [b1, b7, b4, b2], [b6, b3], and [b5, b8]. The PDDL encoding deliberately scrambles the predicates, completely obscuring the tower structure:
\begin{lstlisting}[basicstyle=\ttfamily\small, breaklines=true, frame=single, xleftmargin=0.5em, xrightmargin=2em]
(:init (clear b1)(on b7 b4)(on b5 b8)(clear b6)(ontable b8)(on b4 b2) 
(ontable b3)(ontable b2)(on b1 b7)(handempty)(clear b5) (on b6 b3))
\end{lstlisting}
\vspace{-0.5em}
\noindent Predicates like \texttt{(on b7 b4)} and \texttt{(on b4 b2)}, which belong to the same tower, are separated by unrelated facts such as \texttt{(ontable b8)}, making it impossible to read off the tower structure by sequential scanning. The model must reconstruct the full dependency chains from these scattered facts (see also Appendix~\ref{app:blocksworld_prompt}).

Crucially, just as next-token predictors struggle to identify the correct initial branch in $P^*$ graphs without ``geometric'' insight, a planner lacking global structural awareness will fail to identify the correct tower to unstack among hundreds of distractors. By scaling the number of towers (width) and their heights (depth), we directly test whether the LLM navigates the full $P^*$ structure, correctly identifying the target branch and tracing its dependency chain, or whether it relies on approximate retrieval that degrades with scale. Our analysis of the model's reasoning traces reveals evidence for an active algorithmic simulation strategy; whether latent spatial representations additionally contribute to retrieval efficiency remains an open question.

Much like a human can trivially identify a target block when viewing a physical tower but struggles to visualize an unordered pile of text relations, an LLM might ``see'' or infer the tower geometry by exploiting the rich semantic priors of the Blocksworld domain (e.g., words like ``unstack'' and ``table''). To test whether models rely on this visual-semantic grounding or if they perform genuine structural reasoning, we introduced an isomorphic task by reformulating the Blocksworld scenarios as abstract graph-rewriting problems. This transformation deliberately obscures the domain's tower structure, preventing the LLMs from exploiting preconceived heuristics, such as the necessity to unstack all blocks above a goal node. Additionally, because Blocksworld is a canonical planning testbed, it is likely highly represented in the training corpora. This reformulation ensures the evaluation isolates genuine model-based reasoning rather than remembering a learned task.

Our investigation yields surprising findings that challenge the traditional dominance of classical planners in specific regimes. While satisficing planners like LAMA are designed to trade optimality for speed, our results show that they hit ``computational hard walls'' early along complexity dimensions, particularly when scaling the number of goal blocks. In contrast, we observe that reasoning-enhanced models, specifically Gemini 3.0, do not merely find a solution but frequently converge on the \textbf{theoretical optimal limit}. Rather than hitting a hard computational cliff, these models exhibit a ``zone of instability,'' where they continue to solve highly complex instances optimally even as they begin to fail intermittently.

The contributions of this paper are as follows:

\begin{itemize}[itemsep=1pt, parsep=1pt, topsep=2pt]
    \item \textbf{A Multi-Scenario p-star Problem Space:} We introduce a systematic taxonomy based on p-star structures that isolates distinct complexity axes, including tower height, width, and goal density, alongside a holistic ``Grand Challenge'' that scales all dimensions simultaneously.
    \item \textbf{Structural Analysis via p-stars:} We frame the planning challenge through the lens of p-star structures, analyzing how LLMs manage the retrieval and ordering of dependencies in multi-tower configurations.
    \item \textbf{Optimality Inversion:} We show that as structural complexity (e.g., tower height) scales, a distinct divergence emerges: classical satisficing planners (like LAMA) systematically sacrifice plan optimality to navigate the expanding search space, whereas frontier LLMs can solve significantly more complex instances while preserving theoretical optimality, up until they reach a zone of intermittent failure.
    \item \textbf{Failure Mode Characterization:} We distinguish between the ``hard wall'' failures of classical planners (due to search space explosion) and the ``instability zones'' of LLMs, offering new insights into the reliability and capabilities of LLM-guided planners.
\end{itemize}

\section{Related Work}

The application of Large Language Models (LLMs) to automated planning has evolved from initial skepticism to recent demonstrations of remarkable proficiency using In-Context Learning (ICL).
\paragraph{LLMs and the Planning Debate}
Early investigations painted a pessimistic picture, \citet{valmeekam2023planning} argued on the \textit{PlanBench} dataset that LLMs fundamentally lack the ability to simulate state changes, relying on approximate retrieval rather than true reasoning. \citet{kambhampati2024llms} further posited that while LLMs can support planning (e.g., as heuristic generators), they cannot autonomously generate correct plans due to the absence of an internal world model. Despite these critiques, In-Context Learning (ICL) providing domain descriptions and solved examples has emerged as the dominant paradigm. While earlier models struggled with long-horizon Blocksworld tasks, recent generations show substantially improved performance \citep{valmeekam2023planning,agarwal2024manyshotincontextlearning,bohnet2025enhancingllmplanningcapabilities}.
\paragraph{Frontier Models and Scalability} A significant turning point is the evaluation by \citet{correa2025planning}, assessing GPT-5, Gemini 2.5 Pro, and DeepSeek R1. They showed GPT-5 achieves parity with LAMA on IPC benchmarks, with frontier LLMs solving Blocksworld tasks with up to 477 blocks and plans exceeding 1194 steps. Their use of obfuscated domains revealed that models retain significant symbolic reasoning capabilities even without semantic labels, suggesting ICL engages in logical simulation rather than relying solely on probabilistic priors.
\paragraph{Scaling Test-Time Compute}
\citet{snell2024scaling} demonstrated that dynamically allocating test-time compute based on prompt difficulty can yield better performance than scaling pre-training parameters. While \citet{correa2025planning} established the \textit{feasibility} of large-scale LLM planning, our work shifts the focus to \textit{optimality} and structural analysis via p-star decomposition.

\section{Dataset Generation}
\label{sec:datasets}

We generated a synthetic dataset in the \texttt{blocksworld-4ops} domain \citep{fikes1971strips} to systematically stress-test planning capabilities. Each problem instance is parameterized by a tuple $(W, H, S)$, representing the number of towers, uniform tower height range, and number of target blocks, respectively. 

Secondly, we derive an abstract graph representation for each Blocksworld problem. The objective of this experiment is to mask the underlying physical structure of the tasks (i.e., towers of blocks) to test the model's topological understanding capabilities. In a standard Blocksworld setting, an LLM might solve the problem using memorized semantic priors or heuristics for this class of problem (stacking, constraint satisfaction). Formulating the environment as a graph strips away these semantic hints, preventing the model from relying on common-sense heuristics, such as the need to unstack blocks to free a target block, forcing it to rely entirely on the provided abstract graph formulations' operators and semantics.

To ensure strictly increasing difficulty for Blocksworld and derived graph-problems, we employed rejection sampling during generation. A new instance $P_t$ was accepted only if its optimal plan cost $C(P_t)$ satisfied $C(P_t) \geq C(P_{t-1})$. Target blocks were sampled uniformly from the full height of the towers.

The dataset is organized into four distinct curricula.
\textbf{Search Depth (High Towers):} Evaluates long-horizon reasoning by retrieving a single target block ($S=1$) from fixed-width environments ($W=12$). Tower height $H$ scales linearly from 8 to 1000.
\textbf{Harvest (Goal Blocks):} Evaluates the sequencing of multiple subgoals. The number of goal blocks and number of towers $S$ scales from 4 to 220 in steps of 4 with one goal block per tower. To accommodate increasing targets, width is dynamic while height remains fixed ($H \in [5, 8]$).
\textbf{Interleaved Harvest:} A more challenging variant of Harvest that places two interleaved goal blocks per tower ($S = 2W$), requiring reasoning on intra-branch dependencies. Width scales from 2 to 100.
\textbf{Grand Challenge:} Linearly scales all complexity dimensions simultaneously to test generalization. Parameters range as follows: $W \in [4, 90]$, $H \in [4, 50]$, and $S \in [2, 20]$.

\section{Methodology}

To adapt the general-purpose language model for classical planning, we utilize a strictly formatted one-shot in-context learning (ICL) approach. The model context $\mathcal{C}$ is constructed as a sequence of three components:
\textbf{(1) Domain Specification:} The complete PDDL domain definition, providing the physical rules and action schema.
\textbf{(2) Demonstration Exemplar:} A single tuple $(P_{demo}, \pi_{demo})$ containing a problem instance and its corresponding optimal plan. This demonstration seeds the model with domain logic and enforces strict syntactic adherence to the PDDL plan format.
\textbf{(3) Target Problem:} The query instance $P_{target}$ for which the plan $\pi_{target}$ is generated.

\paragraph{Quantifying Optimality}
We derive the optimal plan length $C_{opt}$ analytically by decomposing the problem into independent \textit{clearing} and \textit{stacking} phases. Let $\mathcal{T}$ be the set of target blocks and $depth(b)$ be the number of blocks covering target $b$. Any covering block is treated as an obstacle and moved to the table ($cost=2$). Once exposed, target blocks are moved directly to the goal configuration; the base target remains stationary, while each subsequent target incurs a construction cost of 2 (pickup and stack). 
For the primary curricula (Harvest, High Towers, Grand Challenge), target blocks were sampled such that no target block was placed in the same tower as another target block, and the base target block was always positioned directly on the table. This constraint simplifies the problem to independent clearing subproblems with a closed-form optimal cost, enabling rigorous optimality analysis. The Interleaved Harvest variant (Section~\ref{sec:harvest}) relaxes this constraint by placing two target blocks per tower, with an adapted analytical cost formula (see Appendix~\ref{app:interleaved_harvest_table}).
The total optimal cost is thus:
\vspace{-0.5em}
\begin{equation}
    C_{opt} = \sum_{b \in \mathcal{T}} 2 \cdot depth(b) + 2(|\mathcal{T}| - 1)
    \label{formula:opt-cost}
\end{equation}
\vspace{-0.5em}

\paragraph{Quantifying Inference-Time Reasoning}
We employ a reasoning-enhanced model variant (Gemini 3.0 Pro) capable of latent "chain-of-thought" generation. Unlike standard greedy decoding, this inference mode allocates a dynamic budget of \textit{reasoning tokens} to facilitate intermediate search and state-tracking prior to generating the final plan. 

To analyze the computational effort allocated to planning, we extract the \texttt{thoughts token count} from the response's \texttt{Metadata}, as exposed by the public Gemini API.\footnote{Documentation on reasoning tokens and usage metadata is available at \url{https://ai.google.dev/gemini-api/docs/thinking}.} This metric allows us to quantify ``inference-time compute,'' i.e., the amount of test-time processing the model dedicates to the problem, and evaluate its correlation with plan validity and optimality.

\subsection{Evaluation and Verification}
Generated plans are validated using the standard \textbf{VAL} validator \citep{howey2004val}. A plan is considered valid only if it is executable from the initial state and satisfies all goal propositions. We report two primary metrics.
\textbf{Success Rate:} The percentage of problems where VAL confirms a valid plan.
\textbf{Optimality Gap:} For valid plans, we compute the cost divergence $\frac{|\pi_{gen}| - C^*}{C^*}$, where $C^*$ is the known optimal cost computed by Formula~\ref{formula:opt-cost}.

\section{Experiments}
\begin{figure*}[th]
    \centering
    \begin{subfigure}[t]{0.48\textwidth}
        \centering
        \includegraphics[width=\textwidth]{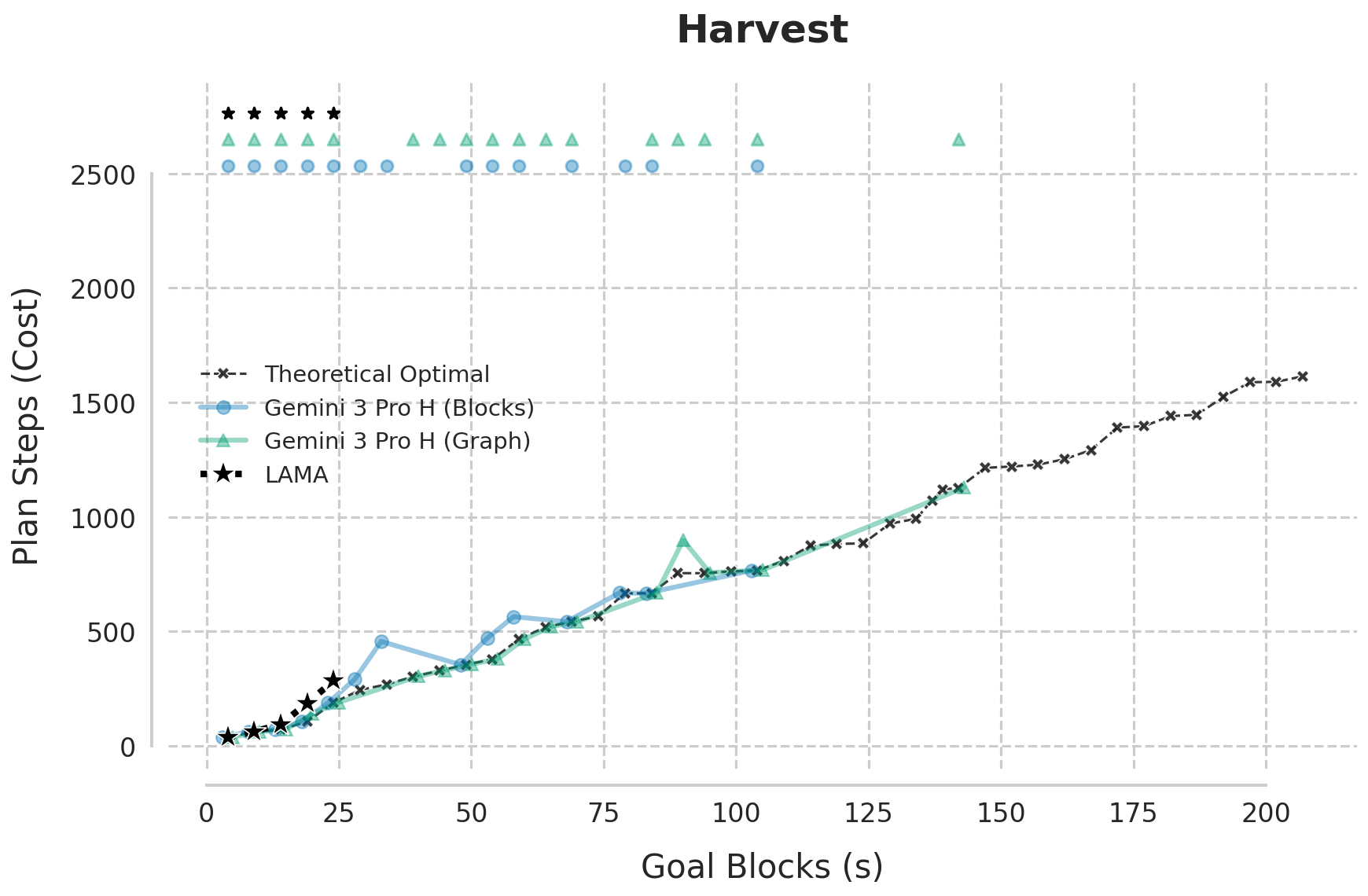}
        \hidecaption\label{fig:goal_per_tower}
    \end{subfigure}
    \hfill
    \begin{subfigure}[t]{0.48\textwidth}
        \centering
        \includegraphics[width=\textwidth]{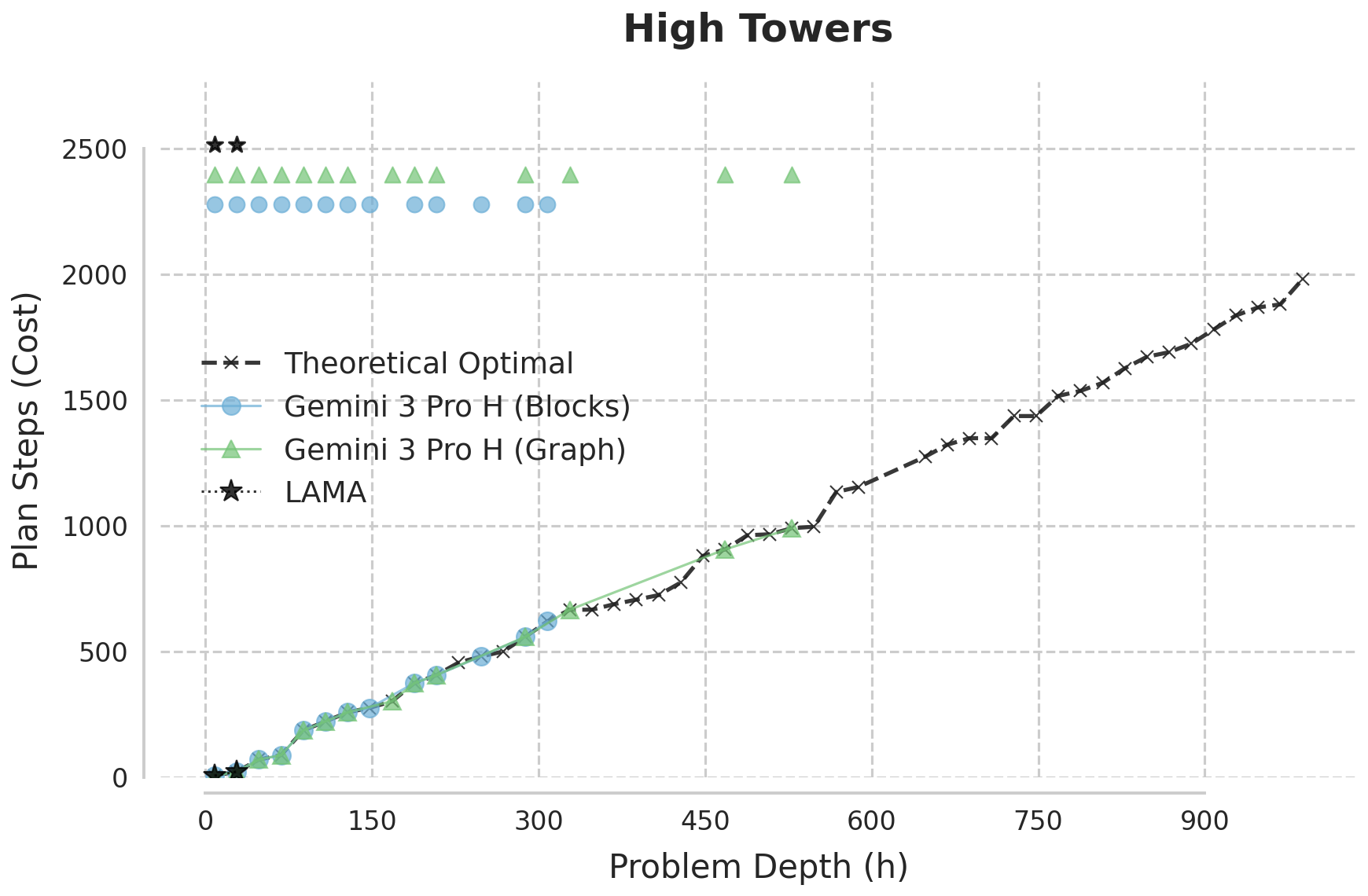}
        \hidecaption\label{fig:high_towers}
    \end{subfigure}
    \vspace{-1em}
    \caption{\textbf{Scaling Goal Blocks and Tower Height.} Gemini 3 Pro vs.\ LAMA-2011 and the theoretical optimal cost. \textbf{(a)} Increasing goal blocks: plan costs illustrating generalization to retrieving and stacking many goal blocks. \textbf{(b)} Increasing tower height: plan costs for retrieving and unstacking from high towers.}
    \label{fig:harvest_and_towers}
\end{figure*}

In this section, we evaluate the planning capabilities and optimality of the models across the curricula defined in Section~\ref{sec:datasets}. To assess both feasibility and plan quality, we plot the number of plan steps (cost) of the generated valid solutions against the problem complexity. These are compared directly to the theoretical optimal cost ($C_{opt}$) computed via Formula~\ref{formula:opt-cost}. As classical baselines, we evaluate both LAMA-2011 \citep{richter2011lama}, a satisficing planner that trades optimality for speed, and A$^*$ with the LM-Cut admissible heuristic \citep{helmert2009landmarks}, which guarantees optimal plans when it terminates. A*/LM-Cut fails on all instances across all four curricula within the 60-second time limit; per-instance results are reported in the Appendix (Tables~\ref{tab:grand_challenge_detailed}--\ref{tab:high_towers_detailed}).

\subsection{Harvest: Scaling Goal Blocks and Towers}
\label{sec:harvest}
Figure~\ref{fig:goal_per_tower} presents the results for the Compositionality experiments (denoted as ``Harvest''), which isolates the model's ability to sequence multiple subgoals by scaling the number of goal blocks ($s$). Both classical planners encounter a hard computational wall at approximately 25 goal blocks: the optimal planner A$^*$/LM-Cut exhausts its search budget, and the satisficing planner LAMA-2011 either fails or produces increasingly suboptimal plans. 

In stark contrast, Gemini 3 Pro evaluated on the standard Blocksworld domain perfectly tracks the theoretical optimal cost line, successfully generating optimal plans for instances with over 100 goal blocks. When evaluated on the abstract graph reformulation (Gemini 3 Pro Graph) the model continues to synthesize optimal plans, demonstrating genuine topological reasoning. However, without the aid of semantic priors (e.g., physical stacking and gravity), the graph formulation exhibits an earlier onset of failures, with intermittent failures beginning around 90 goal blocks and becoming consistent after 150. 
This might also result from the much less common presentation of the problem as a graph-rewrite task. Crucially, when the LLM does succeed, it does not output sub-optimal, heuristic-driven paths; it remains strictly on the theoretical optimal bound.

To probe the limits of the model's decomposition strategy, we also evaluate an \textbf{Interleaved Harvest} variant where each tower contains two interleaved goal blocks ($S = 2W$), breaking the monotonic independence that makes standard Harvest decomposable into trivially independent subproblems. The overall success rate is preserved ($\sim$58\%), matching standard Harvest, but optimality drops sharply: only 10\% of instances are solved optimally (vs.\ 21\% for standard Harvest), with valid plans carrying a consistent $\sim$25--35\% cost overhead. This pattern suggests the model falls back to a less efficient clearing heuristic when intra-branch dependencies prevent the simple ``clear-and-stack'' decomposition, providing evidence that the serial simulation mechanism is \emph{strategy-dependent}: the model has internalized the optimal monotonic strategy but resorts to suboptimal fallbacks when interleaving is required (see Appendix~\ref{app:interleaved_harvest_table} for full results).

\subsection{The Grand Challenge}
\begin{figure*}[th]
    \centering
    \begin{subfigure}[t]{0.48\textwidth}
        \centering
        \includegraphics[width=\textwidth]{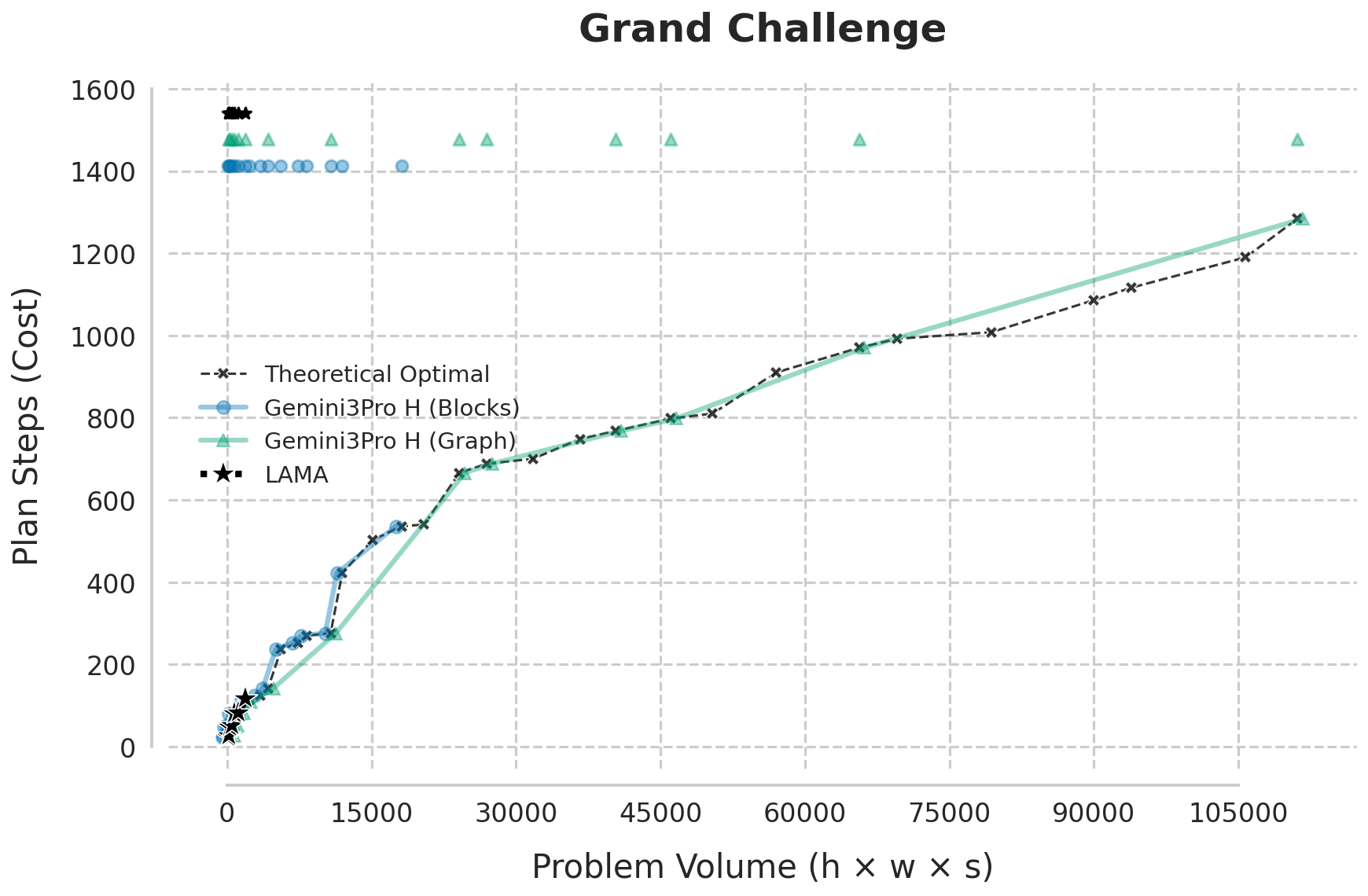}
        \hidecaption\label{fig:compositional_results}
    \end{subfigure}
    \hfill
    \begin{subfigure}[t]{0.48\textwidth}
        \centering
        \includegraphics[width=\textwidth]{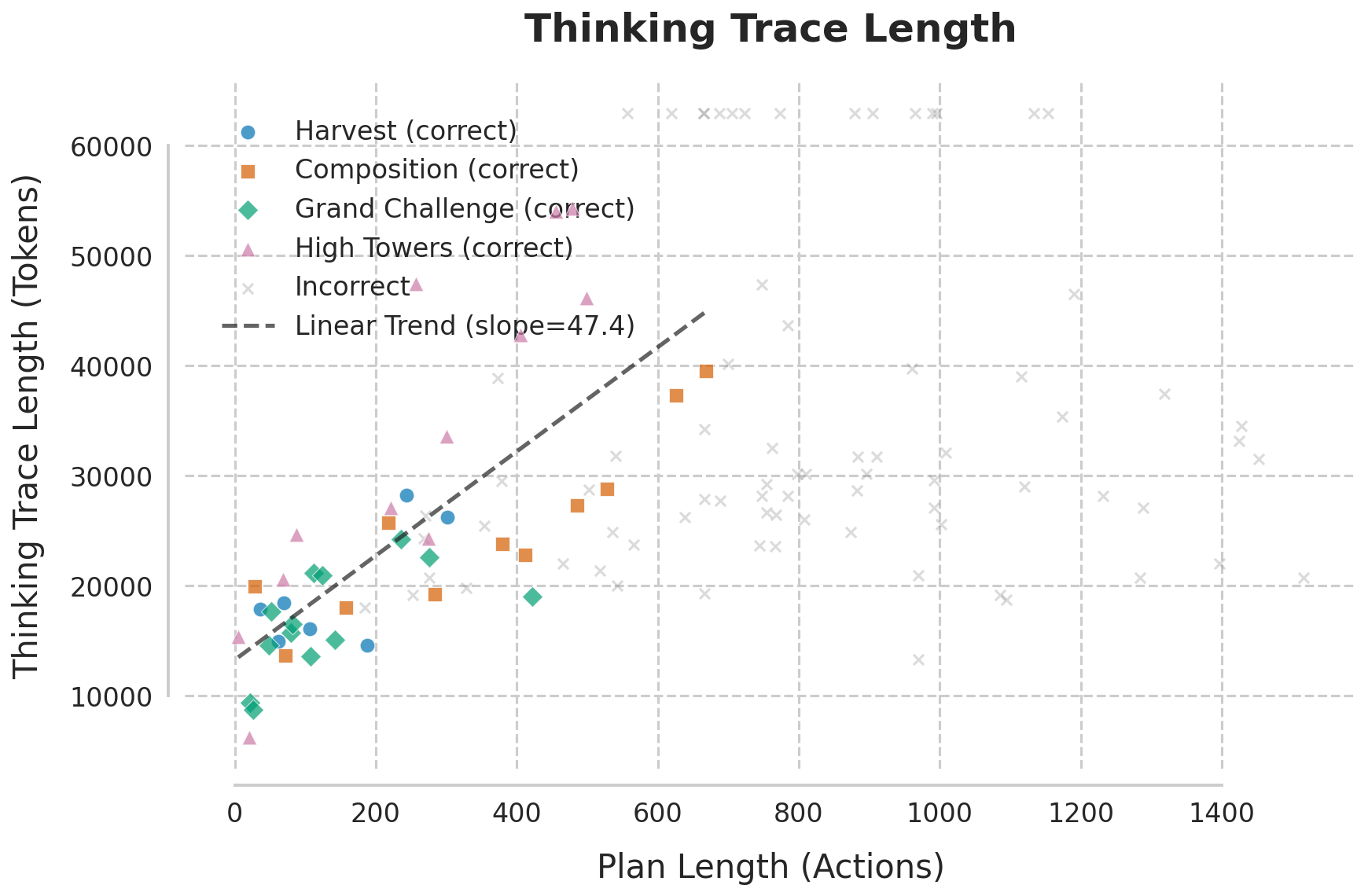}
        \hidecaption\label{fig:inference_scaling}
    \end{subfigure}
    \vspace{-1em}
    \caption{\textbf{Scalability and Compute Allocation.} \textbf{(a)} Grand Challenge: plan steps vs.\ total problem volume ($h \times w \times s$). Gemini 3 Pro tracks optimality into extreme complexities where classical search collapses. \textbf{(b)} Inference-time compute: thinking trace length (tokens) vs.\ optimal cost $C_{opt}$. All curricula follow a linear trend ($\approx$47 tokens/step); failed instances (gray crosses) cluster at higher token counts.}
    \label{fig:grand_and_thinking}
\end{figure*}

To test holistic scalability, the ``Grand Challenge'' experiment scales tower height ($h$), width ($w$), and the number of goals ($s$) simultaneously. As shown in Figure~\ref{fig:compositional_results}, problem volume ($h \times w \times s$) triggers an immediate search space explosion for LAMA, which fails at the lowest volumes tested.

Conversely, the reasoning-enhanced LLM demonstrates extraordinary resilience to combinatorial explosion. Gemini 3 Pro in the Blocksworld domain scales successfully, tracking the optimal cost up to a problem volume of roughly 40,000 before entering a zone of instability where failures become frequent.
Interestingly, the model operating on the obfuscated graph-rewrite domain demonstrates an even deeper maximum scalability in this regime, generating theoretically optimal traversal plans for problem volumes exceeding 100,000. However, it also begins making errors earlier and exhibits more instability at lower volumes compared to the standard Blocksworld formulation. Across both modalities, the transition from success to failure is not characterized by a degradation in plan quality, but rather a binary shift from perfect optimality to complete failure, reinforcing the hypothesis of a strict working-memory or algorithmic simulation limit.

\subsection{Scaling of Inference-Time Compute}
\label{sec:inference_scaling}

The binary transition from perfect optimality to complete failure observed in Figure~\ref{fig:compositional_results} suggests a fixed computational limit. To investigate this, we analyze the allocation of reasoning tokens provided via the Gemini public API interface. Figure \ref{fig:inference_scaling} plots these thinking trace lengths against the optimal plan length $C_{opt}$ across all curricula.

We observe a strikingly consistent linear scaling: for successful plans, the model maintains a stable ratio of approximately 47 tokens per optimal plan step. This high-fidelity relationship supports the hypothesis that the model performs a systematic Serial Algorithmic simulation. The LLM appears to iteratively traverse its internal representation of the problem, where each logical transition in the search space requires a constant, marginal token expenditure. This strict linear relationship aligns with the Serial Scaling Hypothesis \citep{liu2025serial}, which formally shows that sequential decision problems are inherently serial and cannot be shortcut by parallel computation.

Incorrect solutions often cluster near the upper bound of the generated token range, indicating a phase transition where the search budget required for the serial simulation exceeds current generation limits. Notably, the model does not shortcut the task with fewer tokens to pursue a suboptimal plan, but rather maintains this linear compute ratio until it either succeeds optimally or fails entirely.

\section{Discussion}
\label{sec:geometric_p_star}

Having established that frontier models like Gemini 2.5 / 3.0 Pro \citep{comanici2025gemini25pushingfrontier} succeed on these tasks, often utilizing "thought tokens" to overcome the visibility constraint, we now turn to the precise mechanism driving this success. Our analysis centers on Gemini 3.0 Pro, whose API exposes reasoning token metadata essential for the inference-time compute study (Section~\ref{sec:inference_scaling}). While \citet{correa2025planning} demonstrated comparable feasibility across frontier models (GPT-5, DeepSeek R1), our goal is not a model comparison but a structural investigation of \emph{how} and \emph{how well} a reasoning-enhanced LLM navigates the $P^*$ topology. This model-agnostic question remains open, characterized by two competing yet potentially complementary hypotheses:

\textbf{Hypothesis 1: Active Algorithmic Simulation\footnote{We distinguish "Simulation" (tracking the truth values of state predicates like \textit{(clear A)} across time steps) from "Scripting" (creating a fixed sequence of actions like "unstack all"). The model's ability to handle novel, high-depth configurations that require realizing when a previously covered block becomes free suggests a mechanism closer to simulation, though we cannot rule out just repetitive pattern application.}.} The model appears to execute a step-by-step symbolic procedure, (see Appendix \ref{app:thinking_trace}). Evidence for this is found in the thought trace summaries accessible via the Gemini public interface generated by the model. For example, when identifying the need for goal blocks `b3' and `b16', the model often produces a chain of reasoning: ``to get to `b3', I need to clear the stack: `b37', `b14', \dots and `b34'.'' In this view, "thought tokens" function as a working memory that linearizes the search process, effectively running a depth-first traversal algorithm of the tower with the sought goal block to resolve dependencies. However, this hypothesis struggles to explain the retrieval efficiency: how does the model successfully query the correct series of blocks from a shuffled context without an exhaustive $O(N)$ scan?

\textbf{Hypothesis 2: Geometric Navigation.} Alternatively, the model may have a spatial view, navigating a global geometry created when decoding the graph into its kv-cache. As described by \citet{noroozizadeh2025geometric}, deep sequence models can encode dependency chains as spatial clusters, importantly this was only observed when the graphs are stored via SFT, where connected nodes (blocks in a stack) are positioned closer together in the latent space. In this view, the model does not logically deduce every step via brute-force search but rather "senses" the correct path by utilizing the topological proximity of tokens. This geometry effectively creates an $O(1)$ lookup mechanism that solves the hardest step, traversing from the root (table) to the target leaf, without exhaustive search.

\textbf{Summary.} Our available evidence, the structured reasoning traces (see Appendix~\ref{app:thinking_trace} for a full example) and the linear scaling of thinking tokens with plan complexity (Section~\ref{sec:inference_scaling}), strongly supports Hypothesis~1, a \textbf{Serial Algorithmic} mode of operation. In the standard Blocksworld domain, the model explicitly enumerates dependency chains in its thinking trace, identifying which blocks must be unstacked above each goal block and processing them sequentially. These dependencies are often output as structured lists. The linear scaling further corroborates this, as each logical step in the traversal incurs a constant token cost, consistent with systematic multihop reasoning rather than parallel geometric lookup. The model effectively executes a search algorithm within its own context window, utilizing the thought trace as a dynamic memory buffer to maintain state and recursively query the graph. Whether latent geometric representations additionally accelerate retrieval remains an open question requiring access to internal activations. 

Surprisingly, across both Blocksworld and the graph-rewrite problems, the model consistently converges on highly optimal planning strategies. By successfully identifying that it must traverse the isolated ``arms'' of a $P^*$ graph structure, the LLM outperforms both classical satisficing planners (LAMA) and optimal planners (A$^*$/LM-Cut), which fail at the same computational threshold.
It is striking that an LLM can maintain optimal strategy paths at complexities where both optimal search (which guarantees shortest plans but exhausts memory) and satisficing search (which trades optimality for speed) fail entirely. We attribute this to the model's ability to decompose the $P^*$ structure into independent subproblems, where each goal block requires clearing a single branch, and solve them serially via its chain-of-thought. This decomposition effectively reduces an exponentially large joint search space to a linear sequence of manageable multihop lookups, which is precisely the regime where reasoning-token-based simulation excels.

Our isomorphic graph-rewrite reformulation is critical for ruling out reliance on semantic priors. By stripping away the physical semantics of blocks and gravity, we prevent the model from exploiting memorized stacking heuristics. The model's ability to maintain high performance in this obfuscated regime strongly suggests genuine topological reasoning (see Appendix~\ref{app:graph_problem} for the graph-rewrite prompt and model output). The lower graph performance likely stems from the novelty of this framing, to which the model was not exposed in this form in training data, yet the model occasionally solves graph instances that are harder than their Blocksworld equivalents (see Figure~\ref{fig:grand_challenge_h05-10_w006_s02}).

\section{Conclusion}

We moved beyond standard feasibility metrics to evaluate the optimality and structural boundaries of LLMs in autonomous planning. By framing Blocksworld through the topological lens of generalized $P^*$ graphs, we established a problem taxonomy defined by depth, width, and compositionality. Reasoning-enhanced LLMs (Gemini 3.0 Pro) not only solve highly complex configurations that force classical planners into computational cliffs, but track theoretical optimality limits with near-perfect precision. Our obfuscated graph-rewrite formulation demonstrates this is not an artifact of memorization, since even when stripped of domain-specific hints the LLM maintains its ability to navigate the underlying $P^*$ topology, indicating genuine structural reasoning. Analysis of thought traces strongly supports a Serial Algorithmic simulation mechanism, though the remarkable retrieval efficiency suggests a latent Geometric Navigation capability may also contribute.

Our graph-rewriting experiments offer a nuanced perspective on topological reasoning limits. While benchmarks like TopoBench \citep{maniparambil2026topobench} show frontier models struggle with general spatial invariants (connectivity, loop closure), we demonstrate that when a domain is decomposable into specific structural topologies, such as $P^*$ configurations, LLMs successfully navigate graph rewrites without semantic priors. For these specific topologies, frontier LLMs autonomously synthesize optimal strategies that bypass exponential search spaces. Future research should probe internal attention mechanisms to understand efficient branch retrieval, and extend this taxonomy to non-$P^*$ structures.


\section*{Limitations}
Our curricula use controlled goal placement (at most one goal per tower in the standard setting, two in the Interleaved Harvest variant) to enable closed-form optimal cost computation. Extending to arbitrary goal configurations, where multiple goals may require both extraction and relocation within shared towers, remains future work. Additionally, applying the $P^*$ curriculum to other frontier models would clarify whether the observed linear token scaling and binary success-failure transition are universal properties of reasoning-enhanced LLMs.

\clearpage
\bibliography{main}

\clearpage
\appendix
\section{Appendix}
\begingroup
\setlength{\parindent}{0pt} 
\setlength{\parskip}{1em}   

\begin{figure}[ht]
    \centering
    \includegraphics[width=0.6\textwidth]{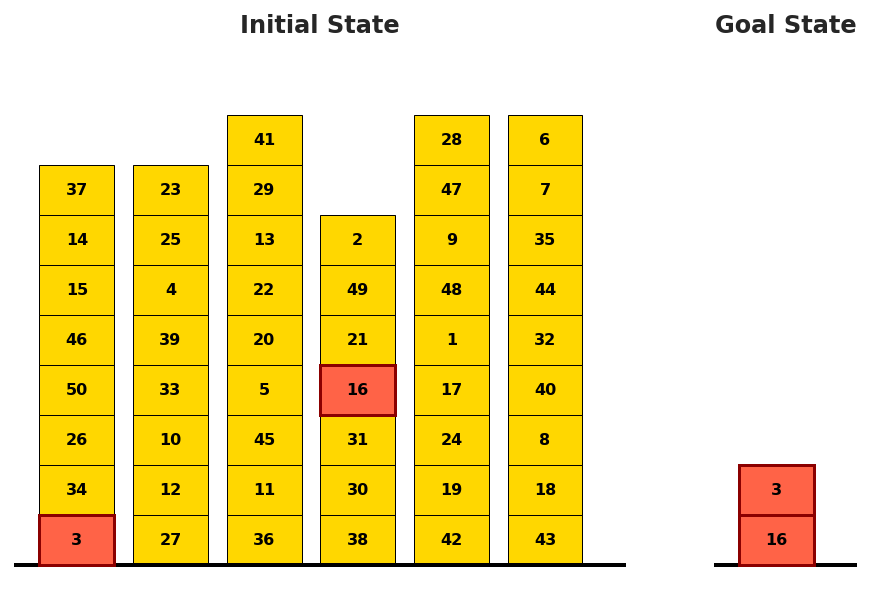}
    \caption{\textbf{Grand Challenge Example.} The simplest problem from the Grand Challenge curriculum with initial state (left) and goal state (right).}
    \label{fig:grand_challenge_h05-10_w006_s02}
\end{figure}

\subsection{Blocksworld Evaluation Prompt}
\label{app:blocksworld_prompt}

To evaluate the optimality of the generated plans, the model was provided with the following prompt. It includes the complete PDDL domain definition for the 4-operator Blocksworld (\texttt{blocksworld-4ops}), strict formatting instructions to ensure zero-shot or few-shot compliance, a solved exemplar, and the final target problem. The selected task is the smallest problem from the Grand Challenge set.
\begin{lstlisting}[basicstyle=\ttfamily\tiny, breaklines=true]

PROMPT:
(define (domain blocksworld-4ops)
  (:requirements :strips)
(:predicates (clear ?x)
             (ontable ?x)
             (handempty)
             (holding ?x)
             (on ?x ?y))

(:action pick-up
  :parameters (?ob)
  :precondition (and (clear ?ob) (ontable ?ob) (handempty))
  :effect (and (holding ?ob) (not (clear ?ob)) (not (ontable ?ob))
               (not (handempty))))

(:action put-down
  :parameters  (?ob)
  :precondition (holding ?ob)
  :effect (and (clear ?ob) (handempty) (ontable ?ob)
               (not (holding ?ob))))

(:action stack
  :parameters  (?ob ?underob)
  :precondition (and (clear ?underob) (holding ?ob))
  :effect (and (handempty) (clear ?ob) (on ?ob ?underob)
               (not (clear ?underob)) (not (holding ?ob))))

(:action unstack
  :parameters  (?ob ?underob)
  :precondition (and (on ?ob ?underob) (clear ?ob) (handempty))
  :effect (and (holding ?ob) (clear ?underob)
               (not (on ?ob ?underob)) (not (clear ?ob)) (not (handempty)))))

Provide only the correct pddl plan. Do not add numbers, items or any additional text. Follow the syntax of the above examplars for the plan if provided. Your plan as plain text:


**Example**

(define (problem BW-rand-4)
(:domain blocksworld-4ops)
(:objects b2 b4 b1 b3)
(:init
(clear b1)
(ontable b2)
(ontable b3)
(clear b3)
(on b1 b4)
(on b4 b2)
(handempty)
)
(:goal (and
(on b2 b1)
(on b4 b3)
))
)

Provide only the result. Do not add numbers, items or any additional text. Follow the format of the above examplars without additional formatting:
(unstack b1 b4)
(put-down b1)
(unstack b4 b2)
(stack b4 b3)
(pick-up b2)
(stack b2 b1)

**TASK**

(define (problem grand_challenge_h05-10_w006_s02)
(:domain blocksworld-4ops)
  (:objects
    b1 b2 b3 b4 b5 b6 b7 b8 b9 b10 b11 b12 b13 b14 b15 b16 b17 b18 b19 b20 b21 b22 b23 b24 b25 b26 b27 b28 b29 b30 b31 b32 b33 b34 b35 b36 b37 b38 b39 b40 b41 b42 b43 b44 b45 b46 b47 b48 b49 b50
  )
  (:init
(on b40 b8)
(on b24 b19)
(on b29 b13)
(on b8 b18)
(clear b2)
(ontable b3)
(on b25 b4)
(on b49 b21)
(on b33 b10)
(on b34 b3)
(on b11 b36)
(on b14 b15)
(on b32 b40)
(clear b41)
(on b47 b9)
(on b22 b20)
(on b6 b7)
(on b15 b46)
(on b45 b11)
(clear b6)
(on b46 b50)
(on b37 b14)
(on b9 b48)
(ontable b43)
(clear b23)
(on b12 b27)
(on b39 b33)
(on b20 b5)
(on b5 b45)
(on b35 b44)
(on b19 b42)
(ontable b38)
(on b4 b39)
(on b17 b24)
(on b21 b16)
(on b28 b47)
(on b10 b12)
(on b50 b26)
(on b48 b1)
(on b18 b43)
(on b13 b22)
(on b23 b25)
(on b7 b35)
(on b41 b29)
(clear b37)
(handempty)
(on b44 b32)
(on b30 b38)
(ontable b42)
(on b16 b31)
(on b2 b49)
(clear b28)
(ontable b27)
(ontable b36)
(on b1 b17)
(on b31 b30)
(on b26 b34)
  )
  (:goal
    (and
      (on b3 b16)
    )
  )
)
Provide only the result. Do not add numbers, items or any additional text. Follow the format of the above exemplars without additional formatting:


\end{lstlisting}

\subsection{Thinking Trace}
\label{app:thinking_trace}

We obtain the following thinking traces via the public interface, which does not include all of the generated thinking tokens. According to the documentation, it is only a high-level summarization (please see the Google Cloud Documentation: \url{https://ai.google.dev/gemini-api/docs/thinking}). Even if it is not a complete thought, it provides valuable insights into the process.

Critically, the second passage below (e.g., \textit{"I now need to start tracing. I use the stack to determine the series of blocks that need to be unstacked."}) indicates algorithmic repeated lookups that it might have carried out before the next high-level summary is provided. In the next summary, passage three, it provides the full list of blocks above: \textit{"I need to clear the stack: `b37`, `b14`, `b15`, `b46`, `b50`, `b26`, and `b34`. Now I have a plan for clearing b3"}. The model has the intuition that these are stacks (towers), probably not because it sees them from the unordered predicates (e.g., \texttt{(on b34 b3)} \texttt{(on b26 b34)}) but from prior knowledge about Blocksworld. Hence we reformulated the blocksworld problem as described in Appendix~\ref{app:graph_problem}.

\begin{lstlisting}[basicstyle=\ttfamily\tiny, breaklines=true]

**PDDL Plan Generation for Blocksworld**

Okay, so I've got this Blocksworld problem, and I need to generate a plan to achieve the goal `(on b3 b16)`.  The domain has the standard Blocksworld operations: `pick-up`, `put-down`, `stack`, and `unstack`. I'm working with objects `b1` through `b50`.

First, I need to analyze the initial state and figure out what needs to move. Looking at the initial state description, I need to determine where `b3` and `b16` are currently located.  I'm going to follow the existing dependencies step by step and work up from what's underneath each block. 
`(on b34 b3)` is stated; thus, I now need to start tracing. I use the stack to determine the series of blocks that need to be unstacked.

So now that I have the dependencies, I can begin constructing sequences of actions to clear the blocks above them. To get to `b3`, I need to clear the stack: `b37`, `b14`, `b15`, `b46`, `b50`, `b26`, and `b34`. Now I have a plan for clearing b3.

For `b16`, tracing dependencies upwards, I see I have blocks `b2`, `b49`, and `b21` on top.\n\nNow I have two sequences.

Now I review the steps to make sure I followed correctly.

I'm ready to move into the execution phase: I'm going to: 
1. Clear `b3`. 2. Clear `b16`. 3. `pick-up b3`. 4. `stack b3 b16`.  Okay, now that I have everything, let's detail the steps.

To clear `b3`:  First, `unstack b37 b14`, then `put-down b37`, and so on. Continue with `b14`, `b15`, `b46`, `b50`, `b26`, and finally `b34`.  At each step, I put-down, then unstack a layer.

To clear `b16`, I must `unstack b2 b49`, `put-down b2`. Then do the same with `b49` and `b21`.

Now `b3` is on the table, and `b16` is clear.

I will `pick-up b3` because it is clear and ontable, and then `stack b3 b16` on top of `b16`.

Let's double check to make sure the plan is optimized.

No other interactions are possible. All seems good. The plan should work.

Double-checking block names and plan against the initial state for consistency, making sure I haven't missed any dependencies or made any typos is important here. So the tower on b3:
`b37, b14, b15, b46, b50, b26, b34, b3` are clear.
The tower on b16 is `b2, b49, b21, b16` with the rest irrelevant.

The plan appears to have no errors. I can start the plan construction now. The final verification is to put the steps into the correct order.

I will ensure to start with:
1. `(unstack b37 b14)`
2. `(put-down b37)`
3. `(unstack b14 b15)`
4. `(put-down b14)`
5. `(unstack b15 b46)`
6. `(put-down b15)`
7. `(unstack b46 b50)`
8. `(put-down b46)`
9. `(unstack b50 b26)`
10. `(put-down b50)`
11. `(unstack b26 b34)`
12. `(put-down b26)`
13. `(unstack b34 b3)`
14. `(put-down b34)`
15. `(unstack b2 b49)`
16. `(put-down b2)`
17. `(unstack b49 b21)`
18. `(put-down b49)`
19. `(unstack b21 b16)`
20. `(put-down b2)`
\end{lstlisting}

\subsection{Graph problem evaluation prompts}
\label{app:graph_problem}

The following prompt, is the rewritten Blocksworld problem.

\begin{lstlisting}[basicstyle=\ttfamily\tiny, breaklines=true]
SYSTEM PROMPT: GRAPH REWRITE SOLVER

I. THE DOMAIN
You are a Graph Rewriting Engine. You operate on a directed graph
representing a hierarchical data structure.

Definitions:
Nodes ($V$):
R: The Root Node (Fixed anchor, infinite outgoing capacity).
T: The Transfer Node (Temporary buffer, capacity = 1 outgoing edge).

Edges ($E$):
Directed edge u -> v represents a parent-child link.
Leaf Node: A node x is a leaf if it has out-degree 0 (it has no children).

II. THE RULES (OPERATIONS)
You can only perform the following four atomic operations to modify the graph topology.

1. DETACH_NODE(child, parent)
Semantics: Detaches a leaf node from its current parent.
(Constraint: Do NOT use this to detach from the Root R).
Preconditions:
- Edge parent -> child exists.
- child is a Leaf (Out-degree = 0).
- T (Transfer Node) is empty (Out-degree = 0).
- parent is not R.
Effect: Delete edge parent -> child. Add edge T -> child.

2. ATTACH_NODE(child, target)
Semantics: Attaches the node currently in the Transfer Node to a new target leaf.
(Constraint: Do NOT use this to attach to the Root R).
Preconditions:
- Edge T -> child exists.
- target is a Leaf node (Out-degree = 0).
- child != target.
- target is not R.
Effect: Delete edge T -> child. Add edge target -> child.

3. ATTACH_TO_ROOT(child)
Semantics: Attaches the node currently in the Transfer Node to the Root R.
Preconditions:
- Edge T -> child exists.
Effect: Delete edge T -> child. Add edge R -> child.

4. DETACH_FROM_ROOT(child)
Semantics: Detaches a leaf node that is currently connected directly to the Root R.
Preconditions:
- Edge R -> child exists.
- child is a Leaf (Out-degree = 0).
- T (Transfer Node) is empty (Out-degree = 0).
Effect: Delete edge R -> child. Add edge T -> child.


Example Input

INITIAL GRAPH STATE (Edges)
R -> n2
R -> n3
n2 -> n4
n4 -> n1

GOAL GRAPH PATTERN (Edges)
n1 -> n2
n3 -> n4

Solution:
DETACH_NODE(n1, n4)
ATTACH_TO_ROOT(n1)
DETACH_NODE(n4, n2)
ATTACH_NODE(n4, n3)
DETACH_FROM_ROOT(n2)
ATTACH_NODE(n2, n1)

Your task:
### INITIAL GRAPH STATE ###
n8 -> n40
n19 -> n24
n13 -> n29
n18 -> n8
Leaf: n2
R -> n3
n4 -> n25
n21 -> n49
n10 -> n33
n3 -> n34
n36 -> n11
n15 -> n14
n40 -> n32
Leaf: n41
n9 -> n47
n20 -> n22
n7 -> n6
n46 -> n15
n11 -> n45
Leaf: n6
n50 -> n46
n14 -> n37
n48 -> n9
R -> n43
Leaf: n23
n27 -> n12
n33 -> n39
n5 -> n20
n45 -> n5
n44 -> n35
n42 -> n19
R -> n38
n39 -> n4
n24 -> n17
n16 -> n21
n47 -> n28
n12 -> n10
n26 -> n50
n1 -> n48
n43 -> n18
n22 -> n13
n25 -> n23
n35 -> n7
n29 -> n41
Leaf: n37
n32 -> n44
n38 -> n30
R -> n42
n31 -> n16
n49 -> n2
Leaf: n28
R -> n27
R -> n36
n17 -> n1
n30 -> n31
n34 -> n26

### GOAL GRAPH PATTERN ###
n16 -> n3
\end{lstlisting}

\begin{lstlisting}[basicstyle=\ttfamily\tiny, breaklines=true]
DETACH_NODE(n2, n49)
ATTACH_TO_ROOT(n2)
DETACH_NODE(n49, n21)
ATTACH_TO_ROOT(n49)
DETACH_NODE(n21, n16)
ATTACH_TO_ROOT(n21)
DETACH_NODE(n37, n14)
ATTACH_TO_ROOT(n37)
DETACH_NODE(n14, n15)
ATTACH_TO_ROOT(n14)
DETACH_NODE(n15, n46)
ATTACH_TO_ROOT(n15)
DETACH_NODE(n46, n50)
ATTACH_TO_ROOT(n46)
DETACH_NODE(n50, n26)
ATTACH_TO_ROOT(n50)
DETACH_NODE(n26, n34)
ATTACH_TO_ROOT(n26)
DETACH_NODE(n34, n3)
ATTACH_TO_ROOT(n34)
DETACH_FROM_ROOT(n3)
ATTACH_NODE(n3, n16)
\end{lstlisting}

The output of the LLM for this simple problem is:

\begin{lstlisting}[basicstyle=\ttfamily\tiny, breaklines=true]
DETACH_NODE(b2, b49)
ATTACH_TO_ROOT(b2)
DETACH_NODE(b49, b21)
ATTACH_TO_ROOT(b49)
DETACH_NODE(b21, b16)
ATTACH_TO_ROOT(b21)
DETACH_NODE(b37, b14)
ATTACH_TO_ROOT(b37)
DETACH_NODE(b14, b15)
ATTACH_TO_ROOT(b14)
DETACH_NODE(b15, b46)
ATTACH_TO_ROOT(b15)
DETACH_NODE(b46, b50)
ATTACH_TO_ROOT(b46)
DETACH_NODE(b50, b26)
ATTACH_TO_ROOT(b50)
DETACH_NODE(b26, b34)
ATTACH_TO_ROOT(b26)
DETACH_NODE(b34, b3)
ATTACH_TO_ROOT(b34)
DETACH_FROM_ROOT(b3)
ATTACH_NODE(b3, b16)
\end{lstlisting}

\subsection{Grand Challenge: Detailed Results}
\label{app:grand_challenge_table}

Table~\ref{tab:grand_challenge_detailed} provides a full breakdown of the Grand Challenge curriculum, which simultaneously scales tower height ($H$), width ($W$), and the number of goal blocks ($S$). We compare LAMA-2011, A*/LM-Cut (both with a 60-second time limit), and Gemini~3~Pro with extended thinking (HIGH) on both the Blocksworld (BW) and isomorphic Graph representations.

\begin{table}[ht]
\centering
\small
\caption{\textbf{Grand Challenge: Detailed Results.} Each row increases all complexity axes simultaneously. ``--'' indicates planner failure within 60s. For Gemini~3~Pro, $\star$ denotes an optimal valid plan matching $C_{\text{opt}}$; $\times$ denotes failure to produce a valid plan. BW = Blocksworld; Graph = isomorphic graph representation.}
\label{tab:grand_challenge_detailed}
\begin{tabular}{rrrrrrrr}
\hline
\textbf{$W$} & \textbf{$S$} & \textbf{$H$} & \textbf{$C_{\text{opt}}$} & \textbf{LAMA} & \textbf{LM-Cut} & \textbf{Gemini 3 Pro (BW)} & \textbf{Gemini 3 Pro (Graph)} \\
\hline
6  & 2  & 5--10  & 22   & 22  & -- & 22   $\star$ & 22   $\times$ \\
9  & 3  & 6--11  & 26   & 26  & -- & 26   $\star$ & 26   $\star$ \\
11 & 3  & 8--13  & 48   & 48  & -- & 48   $\star$ & 48   $\star$ \\
14 & 4  & 9--14  & 52   & 52  & -- & 52   $\star$ & 52   $\star$ \\
17 & 4  & 11--16 & 80   & 80  & -- & 80   $\star$ & 80   $\star$ \\
20 & 5  & 12--17 & 82   & 82  & -- & 82   $\star$ & 82   $\star$ \\
23 & 6  & 14--19 & 108  & 116 & -- & 108  $\star$ & 108  $\star$ \\
\hline
26 & 6  & 15--20 & 112  & --  & -- & 112  $\star$ & 112  $\times$ \\
29 & 7  & 17--22 & 124  & --  & -- & 124  $\star$ & 124  $\times$ \\
32 & 7  & 19--24 & 142  & --  & -- & 142  $\star$ & 142  $\star$ \\
35 & 8  & 20--25 & 236  & --  & -- & 236  $\star$ & 236  $\times$ \\
37 & 9  & 22--27 & 252  & --  & -- & 252  $\star$ & 230  $\times$ \\
40 & 9  & 23--28 & 270  & --  & -- & 270  $\star$ & 244  $\times$ \\
43 & 10 & 25--30 & 276  & --  & -- & 276  $\star$ & 276  $\star$ \\
46 & 10 & 26--31 & 422  & --  & -- & 422  $\star$ & 422  $\times$ \\
49 & 11 & 28--33 & 502  & --  & -- & 502  $\times$ & 502  $\times$ \\
52 & 12 & 29--34 & 536  & --  & -- & 536  $\star$ & 536  $\times$ \\
\hline
55 & 12 & 31--36 & 540  & --  & -- & 540  $\times$ & 540  $\times$ \\
58 & 13 & 32--37 & 666  & --  & -- & 666  $\times$ & 666  $\star$ \\
61 & 13 & 34--39 & 688  & --  & -- & --   $\times$ & 688  $\star$ \\
63 & 14 & 36--41 & 700  & --  & -- & 722  $\times$ & 700  $\times$ \\
66 & 15 & 37--42 & 748  & --  & -- & --   $\times$ & 748  $\times$ \\
69 & 15 & 39--44 & 768  & --  & -- & 710  $\times$ & 768  $\star$ \\
72 & 16 & 40--45 & 798  & --  & -- & --   $\times$ & 798  $\star$ \\
75 & 16 & 42--47 & 810  & --  & -- & 918  $\times$ & 810  $\times$ \\
78 & 17 & 43--48 & 910  & --  & -- & --   $\times$ & 902  $\times$ \\
81 & 18 & 45--50 & 970  & --  & -- & 970  $\times$ & 970  $\star$ \\
84 & 18 & 46--51 & 992  & --  & -- & --   $\times$ & 896  $\times$ \\
87 & 19 & 48--53 & 1008 & --  & -- & --   $\times$ & 962  $\times$ \\
90 & 20 & 50--55 & 1086 & --  & -- & 1072 $\times$ & 1060 $\times$ \\
92 & 20 & 51--56 & 1116 & --  & -- & --   $\times$ & 1092 $\times$ \\
95 & 21 & 53--58 & 1190 & --  & -- & --   $\times$ & 1190 $\times$ \\
98 & 21 & 54--59 & 1284 & --  & -- & 1284 $\times$ & 1284 $\star$ \\
\hline
\end{tabular}
\end{table}

\paragraph{Analysis.}
The Grand Challenge reveals a sharp stratification among planners. A*/LM-Cut fails on all 33~instances, confirming that optimal classical search is intractable at any scale for these multi-goal, multi-tower problems. LAMA-2011 solves the 7~easiest instances ($W \leq 23$, $S \leq 6$) but is already suboptimal at $W=23$ (producing a 116-step plan versus the optimal 108), and fails entirely beyond this point.

Gemini~3~Pro (BW) achieves \emph{perfect optimality} on 16 of 33~problems, with a sharp transition around $C_{\text{opt}} \approx 500$. The Graph representation exhibits a complementary pattern: while it fails on some easy instances where BW succeeds (e.g., $W=6$, $W=26$), it achieves optimal solutions on 14~problems, including several in the failure regime of BW (e.g., $W=58$, $W=69$, $W=72$, $W=81$, $W=98$). This complementary behavior, where Graph succeeds on harder instances that BW cannot solve, mirrors the pattern observed in High Towers and suggests that the graph formulation, while less reliable at low complexities, can unlock deeper structural reasoning into the extreme regime.

\subsection{Harvest: Detailed Results}
\label{app:harvest_table}

The Harvest curriculum isolates width scaling: all towers have fixed height ($H \in [5,8]$) and each tower contributes exactly one goal block ($S = W$). Table~\ref{tab:harvest_detailed} reports performance across 43~instances ranging from 4 to 207~towers.

\begin{table}[ht]
\centering
\small
\caption{\textbf{Harvest: Detailed Results.} Fixed tower height 5--8, one goal block per tower ($S = W$). ``--'' indicates planner failure within 60s. For Gemini~3~Pro, $\star$ denotes an optimal valid plan matching $C_{\text{opt}}$; $\times$ denotes failure to produce a valid plan. BW = Blocksworld; Graph = isomorphic graph representation.}
\label{tab:harvest_detailed}
\begin{tabular}{rrrrrrrr}
\hline
\textbf{$W$} & \textbf{$S$} & \textbf{$C_{\text{opt}}$} & \textbf{LAMA} & \textbf{LM-Cut} & \textbf{Gemini 3 Pro (BW)} & \textbf{Gemini 3 Pro (Graph)} \\
\hline
 4   & 4   & 36   & 36  & -- & 36   $\star$  & 38   $\checkmark$ \\
 9   & 9   & 62   & 62  & -- & 62   $\star$  & 62   $\star$ \\
 14  & 14  & 70   & 94  & -- & 70   $\star$  & 72   $\checkmark$ \\
 19  & 19  & 106  & 186 & -- & 106  $\star$  & 140  $\checkmark$ \\
 24  & 24  & 188  & 284 & -- & 188  $\star$  & 190  $\checkmark$ \\
\hline
 29  & 29  & 244  & --  & -- & 290  $\checkmark$ & 244  $\times$ \\
 34  & 34  & 268  & --  & -- & 456  $\checkmark$ & 268  $\times$ \\
 49  & 49  & 354  & --  & -- & 354  $\star$  & 356  $\checkmark$ \\
 54  & 54  & 378  & --  & -- & 470  $\checkmark$ & 380  $\checkmark$ \\
 59  & 59  & 466  & --  & -- & 564  $\checkmark$ & 466  $\star$ \\
 69  & 69  & 542  & --  & -- & 542  $\star$  & 542  $\star$ \\
 79  & 79  & 666  & --  & -- & 668  $\checkmark$ & 792  $\times$ \\
 84  & 84  & 666  & --  & -- & 666  $\star$  & 668  $\checkmark$ \\
 104 & 104 & 766  & --  & -- & 766  $\star$  & 768  $\checkmark$ \\
\hline
 89  & 89  & 754  & --  & -- & 906  $\times$ & 898  $\checkmark$ \\
 99  & 99  & 762  & --  & -- & 70   $\times$ & 760  $\times$ \\
 114 & 114 & 874  & --  & -- & 1076 $\times$ & 876  $\times$ \\
 134 & 134 & 992  & --  & -- & 176  $\times$ & 1223 $\times$ \\
 167 & 167 & 1292 & --  & -- & 1294 $\times$ & 1092 $\times$ \\
 207 & 207 & 1614 & --  & -- & 16   $\times$ & 1309 $\times$ \\
\hline
\end{tabular}
\end{table}

\paragraph{Analysis.}
The Harvest curriculum reveals a distinctive pattern. LAMA-2011 solves only the 5~easiest instances ($W \leq 24$, $S \leq 24$), and is already heavily suboptimal at $W=14$ (producing 94~steps vs.\ the optimal 70). A*/LM-Cut fails on all 43~instances. Gemini~3~Pro (BW) produces valid plans for 14 of 43~instances, with 9~achieving perfect optimality, including remarkable results at $W=49$ ($C_{\text{opt}}=354$), $W=69$ ($C_{\text{opt}}=542$), and $W=104$ ($C_{\text{opt}}=766$). The Graph formulation exhibits closely correlated performance in this pure-width scaling regime, achieving valid plans on 13 instances (5 optimal). Both representations demonstrate the model's ability to coordinate parallel decomposition across a massive number of independent subproblems at extreme widths. In the failure regime, both models often produce degenerate outputs (e.g., 2 or 16~steps for problems requiring hundreds), suggesting a complete breakdown in plan generation rather than gradual degradation.

\subsection{High Towers: Detailed Results}
\label{app:high_towers_table}

The High Towers curriculum isolates depth scaling: a single forest of 12~towers ($W=12$) with one goal block ($S=1$) is used, while tower height~$H$ ranges from 8 to~993. Table~\ref{tab:high_towers_detailed} reports performance for both the standard Blocksworld (PDDL) and isomorphic Graph representations.

\begin{table}[ht]
\centering
\small
\caption{\textbf{High Towers: Detailed Results.} Fixed width $W{=}12$, one goal block ($S{=}1$), tower height scaled from 8 to 993. ``--'' indicates planner failure within 60s. For Gemini~3~Pro, $\star$ denotes an optimal valid plan matching $C_{\text{opt}}$; $\times$ denotes failure to produce a valid plan. BW = Blocksworld; Graph = isomorphic graph representation.}
\label{tab:high_towers_detailed}
\begin{tabular}{rrrrrr}
\hline
\textbf{$H$} & \textbf{$C_{\text{opt}}$} & \textbf{LAMA} & \textbf{LM-Cut} & \textbf{Gemini 3 Pro (BW)} & \textbf{Gemini 3 Pro (Graph)} \\
\hline
8--13    & 5     & 5   & 5   & 5    $\star$ & 5    $\star$ \\
28--33   & 21    & 21  & --  & 21   $\star$ & 21   $\star$ \\
48--53   & 69    & --  & --  & 69   $\star$ & 69   $\star$ \\
68--73   & 87    & --  & --  & 87   $\star$ & 87   $\star$ \\
88--93   & 185   & --  & --  & 185  $\star$ & 185  $\star$ \\
108--113 & 221   & --  & --  & 221  $\star$ & 221  $\star$ \\
128--133 & 257   & --  & --  & 257  $\star$ & 257  $\star$ \\
\hline
148--153 & 275   & --  & --  & 275  $\star$ & 275  $\times$ \\
168--173 & 301   & --  & --  & 301  $\times$ & 301  $\star$ \\
188--193 & 373   & --  & --  & 373  $\star$ & 373  $\star$ \\
208--213 & 405   & --  & --  & 405  $\star$ & 405  $\star$ \\
228--233 & 455   & --  & --  & 455  $\times$ & 455  $\times$ \\
248--253 & 479   & --  & --  & 479  $\star$ & 477  $\times$ \\
268--273 & 499   & --  & --  & 499  $\times$ & 499  $\times$ \\
288--293 & 557   & --  & --  & 557  $\star$ & 557  $\star$ \\
308--313 & 619   & --  & --  & 619  $\star$ & 619  $\times$ \\
328--333 & 665   & --  & --  & 665  $\times$ & 665  $\star$ \\
\hline
348--353 & 665   & --  & --  & 665  $\times$ & 665  $\times$ \\
368--373 & 687   & --  & --  & 687  $\times$ & 687  $\times$ \\
388--393 & 705   & --  & --  & 703  $\times$ & 705  $\times$ \\
408--413 & 723   & --  & --  & 723  $\times$ & 723  $\times$ \\
428--433 & 773   & --  & --  & 773  $\times$ & 773  $\times$ \\
448--453 & 879   & --  & --  & 161  $\times$ & 879  $\times$ \\
468--473 & 905   & --  & --  & 905  $\times$ & 905  $\star$ \\
488--493 & 961   & --  & --  & 1011 $\times$ & 961  $\times$ \\
508--513 & 965   & --  & --  & 963  $\times$ & 172  $\times$ \\
528--533 & 989   & --  & --  & 503  $\times$ & 989  $\star$ \\
548--553 & 995   & --  & --  & 205  $\times$ & 995  $\times$ \\
568--573 & 1133  & --  & --  & 1131 $\times$ & 1133 $\times$ \\
588--593 & 1153  & --  & --  & 925  $\times$ & 172  $\times$ \\
648--653 & 1273  & --  & --  & 1273 $\times$ & 172  $\times$ \\
668--673 & 1321  & --  & --  & 55   $\times$ & 171  $\times$ \\
688--693 & 1347  & --  & --  & 204  $\times$ & 1    $\times$ \\
\hline
\end{tabular}
\end{table}

\paragraph{Analysis.}
The High Towers curriculum reveals that both representations achieve comparable performance, with the Graph representation solving 14 of 48~problems (29\%) versus 13 for Blocks (27\%). Crucially, both representations achieve \emph{perfect optimality} on every solved instance: all valid plans match $C_{\text{opt}}$ exactly. LAMA-2011 solves only the 2~easiest instances ($H \leq 33$), while A*/LM-Cut fails on all instances. The model's success extends to plans requiring up to 989~steps ($H=528$, Graph) and 619~steps ($H=308$, Blocks). Interestingly, the two representations exhibit complementary strengths: several problems solved by Blocks are failed by Graph and vice versa (e.g., $H=168$ is solved only by Graph; $H=148$ only by Blocks), suggesting partially orthogonal reasoning pathways. In the failure regime at extreme depths ($H > 600$), both representations produce degenerate outputs, truncated plans with costs far below $C_{\text{opt}}$, consistent with a hard inference-time compute ceiling.

\subsection{Interleaved Harvest: Cross-Tower Dependencies}
\label{app:interleaved_harvest}
\label{app:interleaved_harvest_table}

To further probe planning capability, we introduce the \textbf{Interleaved Harvest} curriculum, a variant of the Harvest setup (Section~\ref{sec:harvest}) where each source tower contributes \emph{two} goal blocks instead of one. The goal tower is constructed in round-robin order (deep goals from all towers first, then shallow goals), forcing the planner to alternate between source towers rather than solving each independently.

\paragraph{Optimal Cost.}
The analytical optimal cost is $C_{\text{opt}} = \sum_{i} 2 \cdot d_{\text{deep}}(i) + 2N$, where $d_{\text{deep}}(i)$ is the depth of the deeper goal block in tower $i$ and $N$ is the number of source towers. The formula accounts for the fact that each shallow goal can be directly stacked onto the growing goal tower during clearing (saving 2 actions per tower compared to a table-detour).

\begin{figure}[ht]
    \centering
    \includegraphics[width=0.95\textwidth]{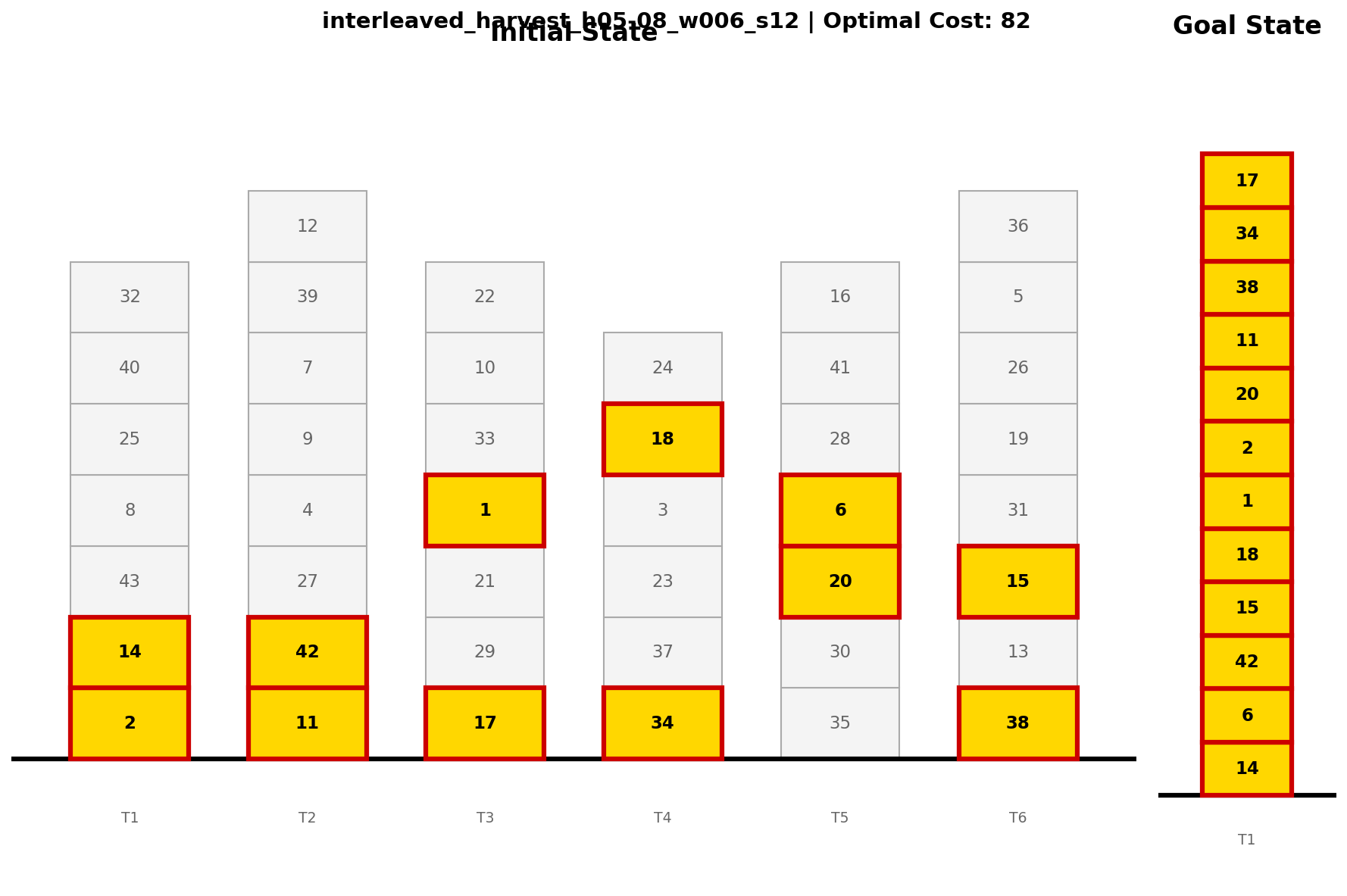}
    \caption{\textbf{Interleaved Harvest Example (6 towers, 12 goal blocks).} Initial state (left) with goal blocks highlighted in gold, and goal tower (right). The scrambled PDDL encoding and cross-tower dependencies make this problem significantly harder for classical planners than the single-goal Harvest.}
    \label{fig:interleaved_harvest_example}
\end{figure}

\paragraph{Results.}
Table~\ref{tab:interleaved_harvest} compares three planners on the Interleaved Harvest problems: LAMA-2011 (satisficing), A*/LM-Cut (optimal), and Gemini~3~Pro with extended thinking (HIGH). All classical planners use a 60-second search time limit. For the LLM, we report the plan length and whether the plan was validated as correct ($\checkmark$) or failed ($\times$).

\begin{table}[ht]
\centering
\small
\caption{\textbf{Interleaved Harvest Results.} Towers with height 5--8, two goal blocks per tower. ``--'' indicates planner failure. For Gemini~3~Pro, $\star$ denotes an optimal valid plan matching $C_{\text{opt}}$; $\checkmark$ denotes a valid but suboptimal plan confirmed by the VAL validator; $\times$ denotes failure to produce a valid plan.}
\label{tab:interleaved_harvest}
\begin{tabular}{rrrrrrl}
\hline
\textbf{Towers} & \textbf{Goals} & \textbf{$C_{\text{opt}}$} & \textbf{LAMA} & \textbf{LM-Cut} & \textbf{Gemini 3 Pro} & \\
\hline
2  & 4   & 26  & 26  & -- & 26  & $\star$ \\
4  & 8   & 38  & 38  & -- & 38  & $\star$ \\
6  & 12  & 82  & 82  & -- & 82  & $\star$ \\
8  & 16  & 84  & 84  & -- & 110 & $\checkmark$ \\
10 & 20  & 106 & 106 & -- & 132 & $\checkmark$ \\
12 & 24  & 152 & 260 & -- & 182 & $\checkmark$ \\
14 & 28  & 160 & 312 & -- & 200 & $\checkmark$ \\
16 & 32  & 190 & 326 & -- & 236 & $\checkmark$ \\
18 & 36  & 196 & 576 & -- & 286 & $\checkmark$ \\
20 & 40  & 238 & 614 & -- & 302 & $\checkmark$ \\
\hline
22 & 44  & 262 & -- & -- & 334 & $\times$ \\
24 & 48  & 264 & -- & -- & 356 & $\checkmark$ \\
26 & 52  & 304 & -- & -- & 404 & $\checkmark$ \\
28 & 56  & 308 & -- & -- & 308 & $\star$ \\
30 & 60  & 332 & -- & -- & 436 & $\times$ \\
32 & 64  & 358 & -- & -- & 462 & $\checkmark$ \\
34 & 68  & 390 & -- & -- & 510 & $\checkmark$ \\
38 & 76  & 440 & -- & -- & 586 & $\checkmark$ \\
72 & 144 & 788 & -- & -- & 788 & $\star$ \\
78 & 156 & 860 & -- & -- & 1122& $\checkmark$ \\
\hline
\end{tabular}
\end{table}

\paragraph{Analysis.}
The results reveal a striking contrast between classical planners and the LLM on this cross-tower dependency task. A*/LM-Cut, the optimal classical planner, fails on \emph{all} 50 instances, including the smallest with just 2~towers and 4~goal blocks, demonstrating that the interleaved goal structure renders exhaustive optimal search intractable even at minimal scale. LAMA-2011 solves problems with up to 20~towers (40~goal blocks) but produces increasingly suboptimal plans before failing entirely at 22~towers.

Gemini~3~Pro produces valid plans for 18 of the 50~instances, including 8~problems beyond LAMA's failure threshold (e.g., 24, 26, 28, 32, 34, 38, 72, and 78~towers). On the first three problems (2--6~towers, 4--12~goal blocks), the LLM achieves \emph{optimal} plan length, matching $C_{\text{opt}}$ exactly. Remarkably, the model also produces optimal plans at 28~towers (56~goals, $C_{\text{opt}}=308$) and 72~towers (144~goals, $C_{\text{opt}}=788$). Where LAMA does find plans, it is consistently less efficient than the LLM: for instance, at 12~towers LAMA produces a 260-step plan while the LLM achieves 182~steps (optimal: 152). These results extend the findings of the main paper to a setting with explicit cross-tower dependencies, further supporting the hypothesis that the model performs structured reasoning over scrambled relational inputs rather than relying on shallow pattern matching.

\endgroup

\end{document}